\documentclass[letterpaper, 10 pt, conference]{ieeeconf}  

\IEEEoverridecommandlockouts                              
\usepackage{booktabs}
\usepackage{todonotes}
\usepackage{subfigure}
\usepackage[noadjust]{cite}
\usepackage{algorithm}
\usepackage{algorithmic}
\usepackage{mcode}
\usepackage[hidelinks]{hyperref}
\usepackage{siunitx}
\usepackage{caption}
\title{\LARGE \bf FastOrient: Lightweight Computer Vision for Wrist Control in Assistive Robotic Grasping*}

\author{Mireia Ruiz Maym\'o, Ali Shafti, and A. Aldo Faisal
\thanks{*Research supported by eNHANCE (\href{http://www.enhance-motion.eu}{http://www.enhance-motion.eu}) under the European Union's Horizon 2020 research and innovation programme grant agreement No. 644000.}
\thanks{M. R. Maym\'o, A. Shafti and A. A. Faisal are with the Brain and Behaviour Lab, Dept. of Computing and Dept. of Bioengineering, Imperial College London, SW7 2AZ, London, UK.
        {\tt\small a.shafti, a.faisal@imperial.ac.uk}}
}

\begin{document}

\maketitle
\thispagestyle{empty}
\pagestyle{empty}

\begin{abstract}
Wearable and Assistive robotics for human grasp support are broadly either tele-operated robotic arms or act through orthotic control of a paralyzed user's hand. Such devices require correct orientation for successful and efficient grasping. In many human-robot assistive settings, the end-user is required to explicitly control the many degrees of freedom making effective or efficient control problematic. Here we are demonstrating the off-loading of low-level control of assistive robotics and active orthotics, through automatic end-effector orientation control for grasping. This paper describes a compact algorithm implementing fast computer vision techniques to obtain the orientation of the target object to be grasped, by segmenting the images acquired with a camera positioned on top of the end-effector of the robotic device. The rotation needed that optimises grasping is directly computed from the object's orientation. The algorithm has been evaluated in 6 different scene backgrounds and end-effector approaches to 26 different objects. 94.8\% of the objects were detected in all backgrounds. Grasping of the object was achieved in 91.1\% of the cases and has been evaluated with a robot simulator confirming the performance of the algorithm.

\end{abstract}
\section{Introduction}
The human upper limb enables reaching, grasping and manipulations. Patients with paralysis, be it e.g. due to spinal cord injuries or neurodegenerative diseases, or amputees suffer from severe actuation limitations in daily life, leading to difficulties in reaching and/or grasping. Assistive robotic devices help in these cases, providing support to the patient's now limited actuation capabilities. These devices, however, require low-level control by the end-user, either through residual motion (e.g. head controlled computer mouse \cite{sim2013head}) or through neural interfaces \cite{fara2013robust,farina2016reflections}. The former is not always feasible, depending on the level of paralysis or amputation, and the latter is limited by the number of independent channels that can be read out (typically outnumbered by the degrees of freedom within wearable or assistive robotic devices). We need solutions to obtain and decode human action  intentions from other sources, and feed them into the control loop of the robotic device for intuitive and natural interaction.

\begin{figure}[tp]
\includegraphics[width=\columnwidth]{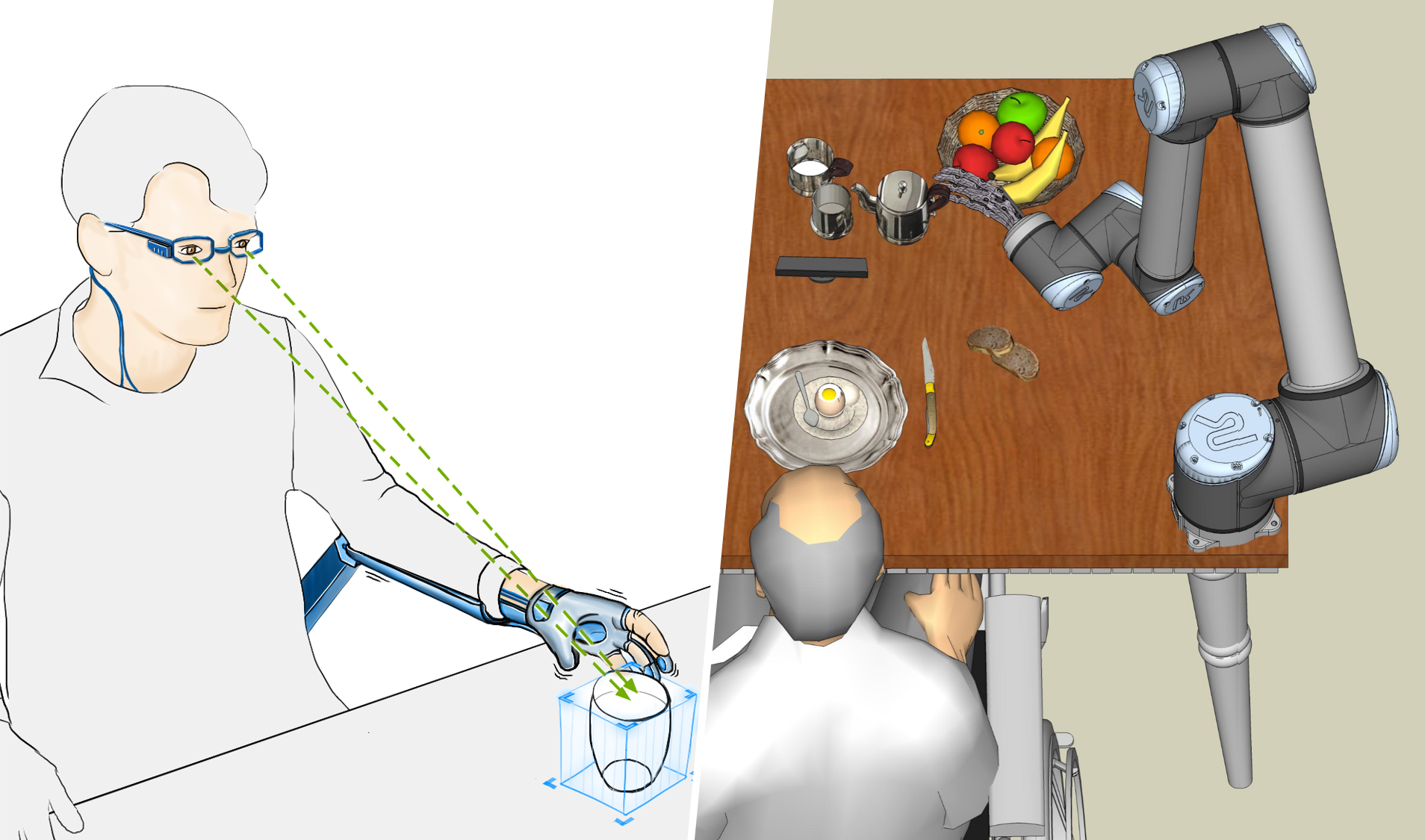}
\caption{Concepts of assistive robotic devices for reaching and grasping. Left: Robotic orthotic glove. Right: Tele-operated robotic hand. Both require wrist rotation control for suitable grasping.}
\label{concept}
\end{figure}

Different modalities have been used for this, with the aim to enhance the way such devices interact with human users. Multi-modal systems relying on unaffected abilities, e.g. the use of gaze-based robotic end-point control for reaching assistance \cite{Tostado20163DActuators,Dziemian2016Gaze-basedDrawing,Maimon-Dror2017TowardsTracking}, or the use wearable robotics controlled through eye winks and  voice \cite{Noronha2017WinkGloves}. In all these cases, a suitable eventual grasp is only possible with the correct orientation of the hand; be it the human hand using an orthotic (Figure \ref{concept}-left), or a tele-operated robotic hand, performing the grasp (Figure \ref{concept}-right).

A third option is to use the context of the user and the robotic actuator to infer the most likely action intention, be it from a dictionary of human activity \cite{thomik2013real} or the dynamics of the preceding movements \cite{xiloyannis2017gaussian}. This paper describes \emph{FastOrient}, a compact computer vision-based algorithm for the control of a robotic wrist's rotation to obtain a suitable grasp of the object of interest. The FastOrient algorithm effectively off-loads complicated low-level details of grasping orientation to sub-systems that provide automated orientation adjustment, thereby reducing cognitive load and training time for the user, and thus ultimately user uptake and technology embodiment
\cite{makin2017neurocognitive}.
\section{Background}
Camera based approaches for end-effector orientation control have been deployed in robotics. Typically, a camera positioned on top of the end-effector or on the user's head to detect certain characteristics of the target object. These include multi-camera approaches that segment and identify spherical and cylindrical objects in a fixed workspace\cite{McMullen2014DemonstrationProsthetic}: they combine  an eye-tracking device to detect the user's gaze point and identify the target object and an  RGBD camera  positioned in a fixed mount that allows extracting depth and RGB information from the environment. Final robotic hand position and grasp depends on object shape and  orientation. Similarly in \cite{Sanz1998Vision-guidedRobots}, where end-effector camera and object segmentation are used for object boundary extraction and detection of orientation. In \cite{Markovic2015SensorProsthesis}, the shape, size and orientation of target objects are estimated using a camera located at the user's head level and the algorithm extracts features by approximating the object with geometrical models. In \cite{Dosen2010CognitiveEvaluation} computer vision techniques are developed to locate target objects and estimate their size and orientation using a camera and an ultrasound sensor. Wrist rotation control using image segmentation of a grayscale filtered image was added to this system \cite{Dosen2011TransradialPrehension}. In autonomous robotics \cite{Saxena2008RoboticVision} proposed a machine-learned  algorithm that identifies  points for grasping using two or more images of the same object acquired by an end-effector camera. The more general pixel-to-torque problem (using cameras to control robots) is currently only feasible in very large scale simulation settings of robot grasping and vision using deep learning \cite{Redmon2015,Bousmalis2018}. However, these approaches currently lack interpretability and may not translate well to safety-critical human-in-the-loop and wearable robotics settings. 

It has been shown by \cite{Cutkosky1989OnTasks,Feix2016TheTypes} that humans tend to grasp objects along their long axis. This grasp allows taking the object with high stability and enables correct grasping in most cases. Thus, grasping orientation is not important for spherical or rounded objects, as all their axes are equal. Our assumption is that reaches often terminate in grasps and the object's that we can obtain the correct orientation of the wrist from the principal axes of the object. This forms the basis for the method described in this paper, which aims to achieve suitable grasping through a computationally light-weight and compact algorithm.
\section{Methodology}
\begin{figure}[tp]
\includegraphics[width=\columnwidth]{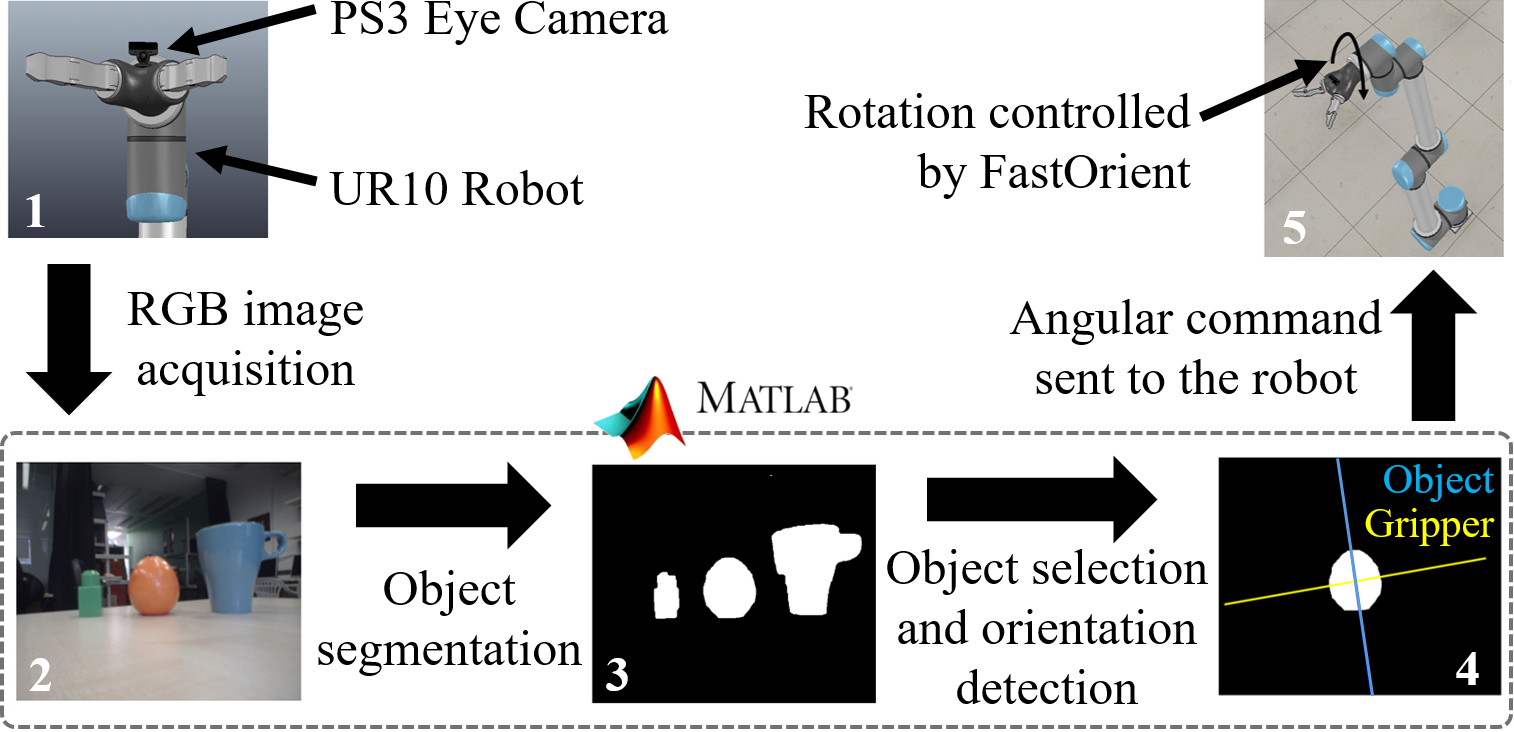}
\caption{General architecture of FastOrient. (1) Camera on top of the UR10 gripper; (2) RGB image during approaching movement; (3) Segmented image with detected objects; (4) Target object identification and orientation computed; (5) UR10 Robot in VRep simulator environment.}
\label{arch}
\end{figure}

Wrist orientation control is the key step to providing a suitable grasp of an object of interest. The general architecture of our algorithm, titled \emph{FastOrient}, is presented in Figure \ref{arch}. First, an RGB image is acquired with a PlayStation 3 Eye Camera (PS3 Eye). Then, different image processing techniques are applied to the image to detect and identify the target object of interest. The detected object's orientation is then computed with reference to the corresponding gripper's orientation. This orientation is sent to the robot controller to rotate its end-effector and allow a suitable grasp of the object. For the purpose of this paper, a Universal Robots UR10 arm is used in simulation - this is interchangeable with any robotic system that allows wrist rotation.

\emph{FastOrient} is implemented using MATLAB (R2015b 32-bit) with functions from its Image Processing Toolbox. VRep (VRep PRO EDU 3.3.1 32bit Version), a robot simulation environment, is used to reproduce and evaluate the algorithm on a robotic system. The general structure of \emph{FastOrient} can be seen in Algorithm \ref{alg1}. Note that 15 frames are obtained for analysis, this is to increase robustness by considering the median orientation, avoiding effects arising from potentially deviated detections.
\begin{algorithm} [bp]
\caption{FastOrient}
\begin{algorithmic} [1]
            \STATE Initialise camera;
            \FOR{$i=0$ \TO $15$} 
            \STATE Acquire RGB image;
            \STATE Paint grayscale pixels green;
            \STATE Project image into HSV colourspace;
            \STATE Select target object via minimum distance algorithm;
            \STATE Extract features;
            \STATE Calculate object orientation;
            \ENDFOR
            \STATE Calculate median orientation;
            \STATE Calculate suitable gripper orientation;
            \RETURN estimated angle to robot;
\end{algorithmic}
\label{alg1}
\end{algorithm}
\subsubsection{Image acquisition} The PS3 Eye is a digital camera producing 640x480 pixel images at a 60Hz frame rate. It integrates an infrared filter and two adjustable fixed-focus zoom lenses. The 75-degree field of view lens was selected to enable acquisition of planes at long distances and avoid blurred objects within the image. The camera is positioned on top of the end-effector of the device. This is a more intuitive positioning than those proposed by \cite{Markovic2015SensorProsthesis,McMullen2014DemonstrationProsthetic} as it allows to get the "hand's point of view" during reaching movement. This positioning is more affected by motion noise but ensures the object of interest is located in the camera's field of view.
\subsubsection{Color spaces} Shadows can be identified as parts of the object itself and illumination spots may impair the complete detection of the item of interest. \cite{Zhao2002RobustSpace} reports that shadows share chromaticity values with the background but the brightness of pixels corresponding to these shaded areas is considerably lower. Accordingly, the Hue-Saturation-Value (HSV) space is chosen here to account for these differences. The Red-Green-Blue (RGB) model cannot separate chromaticity from brightness, while in HSV, the \emph{Value} channel represents the brightness and, the \emph{Hue} and \emph{Saturation} channels together give rise to chromaticity \cite{Zhao2002RobustSpace}. Different thresholds can be directly applied to the \emph{Saturation} and the \emph{Value} channels for the removal of shadows and illumination effects. The application of thresholds in the HSV channels removes the effects of shadows but also limits the detection of grayscale objects. As a result, pre-processing of the image in the RGB space is needed. The grayscale parts of the image are thus painted in a different color, in RGB and before projection to HSV, so that they can be easily detected later. The algorithm counts the number of black (RGB code: $[0, 0, 0]$) or dark gray (RGB code range considered: $[95$-$125, 95$-$125, 95$-$125]$) pixels of the acquired RGB image (see Figure \ref{orientation}.1). This is done to distinguish the cases in which black or dark backgrounds occupy a large part of the image. If this is the case, the pixel colors should remain unchanged so as to avoid the background being selected as the object. Accordingly, if more than $100,000$ pixels are considered dark (the acquired images have $307,200$ pixels, i.e. $32.5\%$), the algorithm does not change their color. However, it will look for white pixels and change their color to green, so that white objects can be detected in dark backgrounds. Nevertheless, if the number of dark pixels is below the threshold, the algorithm paints different ranges of grayscale pixels: black, gray and white, in green color. This is done by directly changing the colour of the pixel to the RGB value $[77, 153, 77]$, which corresponds to green, as seen in Figure \ref{orientation}.2. Finally, a new RGB image is built with the corresponding color modifications so that the image segmentation can start with all the information concerning grayscale objects.

\subsubsection{Image segmentation}The main goal of the image segmentation procedure is to convert the image into information easier to analyse.

\begin{figure}[tp]
\includegraphics[width=\columnwidth]{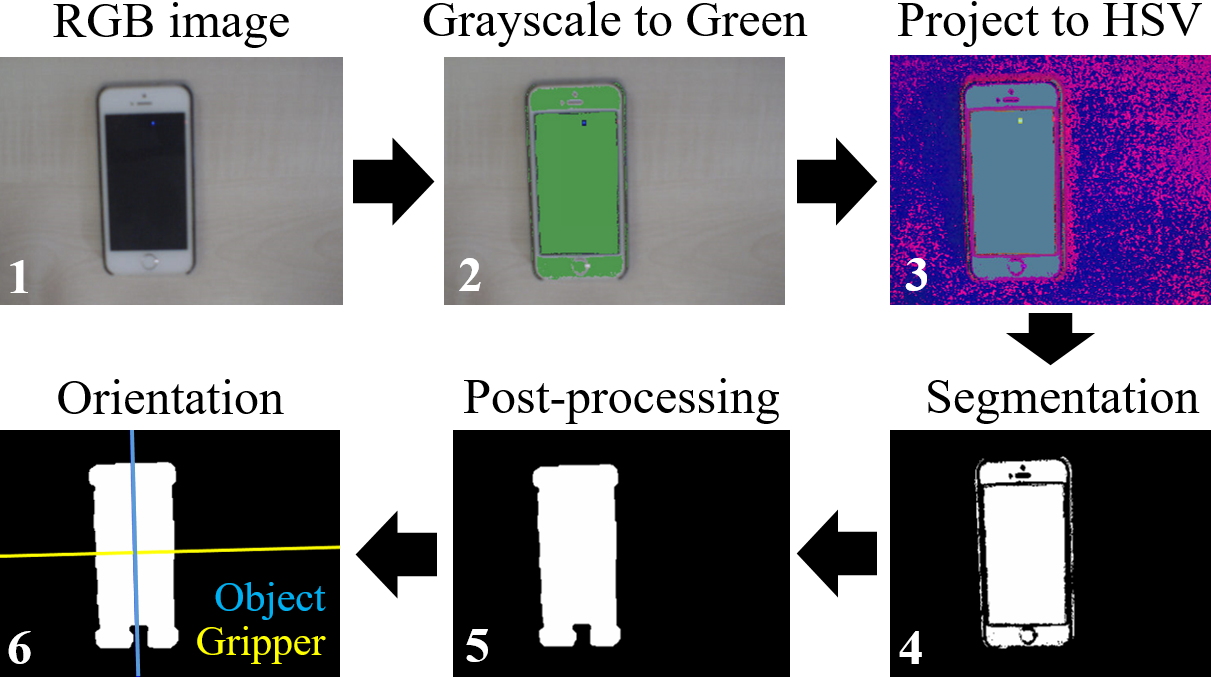}
\caption{Computer vision pipeline for complete target detection procedure. (1) RGB camera image, (2) Grayscale pixels detected and painted in green, (3) Projection into  HSV color space, (4) Image segmentation for analysis of object's features, (5) Post-processing to improve detection quality, (6) Object's orientation (blue) and gripper's orientation (yellow) calculated.}
\label{orientation}
\end{figure}

\emph{Binarisation}: As seen in Figure \ref{orientation}.4, this consists of the transformation of the original image into a black and white one. In the binary image, the object pixels are white and the background is represented in black. This conversion is performed by applying a threshold on the HSV values to separate the object's pixels from the background ones considering each channel's histogram. Otsu's method is applied to find the optimal threshold for each image. This method assumes the image contains two classes of pixels, foreground and background, and considers the gray-level histogram of each class distribution. Then, it defines the optimal threshold that minimizes the overlap between both distributions and allows a better discrimination between the classes \cite{Jianzhuang1991AutomaticMethod,Otsu1979AHistograms}. Better results were observed if only the \emph{Saturation} and the \emph{Value} channels were thresholded and \emph{Hue} information was not considered for image segmentation. Finally, the information of both channels is combined using an intersection (logical \mcode{and} operation).

\emph{Post-processing of the Binarised Image}: In order to improve the quality of object detection and make the detected shape more precise, different post-processing methods are applied (see Figure \ref{orientation}.5). Areas smaller than a threshold of 1500 pixels (value found empirically) are removed with an opening function. This is done to remove all the small areas of the detection that do not correspond to the object but to background or shadows. Holes in the detected parts are filled using the \mcode{imfill} MATLAB function. A combination of opening and closing functions are also applied to remove noise and make shapes smoother using the \mcode{imopen} and \mcode{imclose} MATLAB functions. 
\subsubsection{Feature Extraction} The different properties from the objects detected in the image have been measured using the \mcode{regionprops} MATLAB function. The features extracted from the objects are as follows. The centroid: \mcode{regionprops} calculates the center of mass of the region used to calculate the minimum distance algorithm. The bounding box: \mcode{regionprops} gives the smaller rectangle containing the region used to create a mask for target object discrimination. The orientation: \mcode{regionprops} returns a scalar corresponding to the angle of the long axis of an ellipse that has the same second moment of inertia as the object region, with respect to the horizon.
\subsubsection{Target object detection} To discriminate the target object from all the other objects that are detected in the image, the same procedure as \cite{Dosen2010CognitiveEvaluation} is used, selecting the object located closest to the center of the image.
\subsubsection{Robotic end-effector orientation calculation}
The orientation of the detected object is calculated for each obtained frame and the median is computed. Then, the perpendicular orientation is calculated for the end-effector orientation in order to place the gripper along the long axis of the object (see Figure \ref{orientation}.6). This orientation is sent as a command to the robotic arm as the output of \emph{FastOrient}.
\begin{figure}[!htb]
\includegraphics[width=\columnwidth]{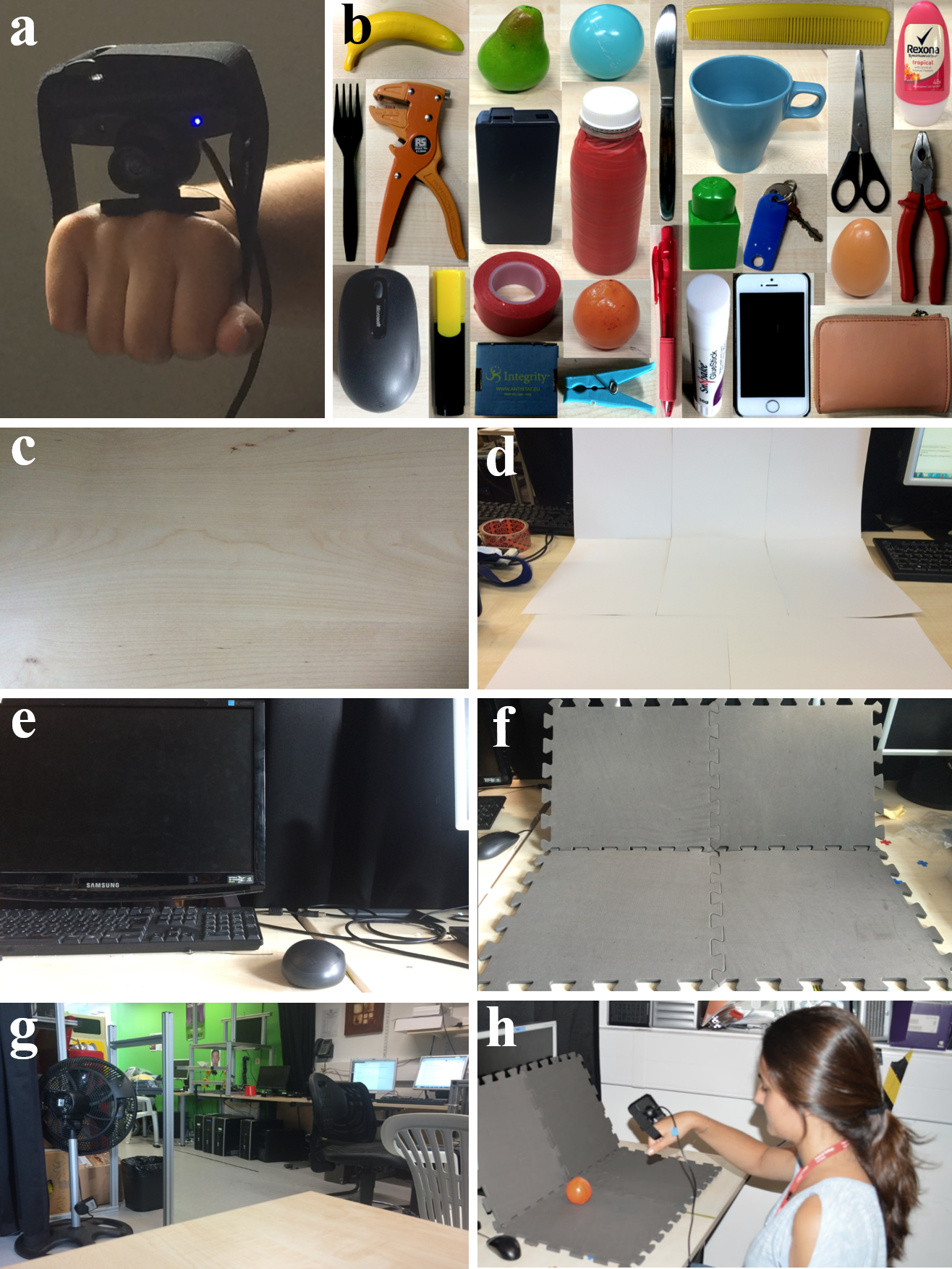}
\caption{Evaluation experiment setup and definitions. (a) Camera placement on operator's hand to collect image data from natural human reaches, (b) our 26 objects used for evaluation,  various backgrounds and surfaces used for evaluating robustness: (c) Beige table background, (d) White background, (e) Dark background, (f) Desk background, (g) Open lab background, (h) Evaluation setup with the operator reaching for the object of interest.}
\label{experiment}
\end{figure}
\section{Evaluation experiments and results}
The evaluation is divided into three sections: (1) detection of the target object, (2) the object and gripper orientation as returned by \emph{FastOrient}, and (3) the rotation of the robotic gripper in VRep, the robotic simulation environment. The evaluation is performed in six different experiment setups considering 26 different objects with each approach repeated ten times.
\subsection{General conditions}
To simulate the reaching movement, the PS3 Eye is positioned on top of an operator's right closed fist to collect image data from natural human reaches. The base of the camera is in contact with the back of their hand and the lens is focusing forwards (see Figure \ref{experiment}.a). A total of 26 different objects are tested for the evaluation of the control algorithm (see Figure \ref{experiment}.b). These are objects used in activities of daily living, with different shapes, colors and materials. Five different backgrounds are defined to evaluate the system. These backgrounds represent almost all possible daily situations for object grasping (see Figure \ref{experiment}.c-g). The lights in the lab are on during the experiments and the curtains are open in order to have natural light.

Reaching approaches are divided into four types, (1) Top approach: is defined as being perpendicular to the surface where the object is positioned; (2) Lateral approach: is parallel to this surface; (3) Angulated approach: is at approximately 60 degrees with respect to the surface, and (4) Variable approach: represents the most natural approach for each object, as preferred by the human operator. The arm trajectory follows a straight line from its initial position (approximately 35cm from the object) to its final point (approximately 20cm from the object). Combinations of the reaching approach types and backgrounds are then selected as follows to define six experiment setups:
1. Beige Table + Top, 2. White + Top/Lateral (depending on the object), 3. Desk + Lateral, 4. Dark + Top/Lateral (depending on the object), 5. Open Lab + Angulated (Open Lab 1), and 6.  Open Lab + Variable (Open Lab 2).
\subsection{Evaluation of target object selection}
Target object selection is evaluated for all objects and backgrounds. In order to assess performance, the output image obtained after the segmentation process is classified through visual inspection, into categories as follows. \emph{Complete detection}: output shape corresponds exactly to the real shape of the object; \emph{Partial detection affected by shadows}: output shape is not the same as the real one, some parts are added due to effects of shadows; \emph{Partial detection affected by illumination}: output shape is incomplete, some parts are missing due to effects of illumination; \emph{Partial detection affected by gray parts}: output shape is incomplete, all the parts are detected except for the grayscale parts in the object; \emph{None}: the object is not detected at all.
\subsection{Evaluation of object orientation detection}
The output of the object orientation detection, shown visually as a line drawn along the object's axis as detected on the image, is also evaluated through visual inspection. The returned orientation is then classified as follows. \emph{Complete alignment} with the object's long axis; \emph{Affected by partial detection} where the detected orientation deviates from the long axis by $<\ang{20}$, grasping is still allowed; \emph{Incorrect} where deviation is $>\ang{20}$, grasping is not possible; \emph{None}, where no orientation is retrieved due to the object not being detected in the previous step.
\subsection{Evaluation with robot simulation}
VRep is an environment for robot simulations. The robot used for the simulation is the Universal Robots UR10 robot, an anthropomorphic 6 degrees-of-freedom robotic arm, equipped with a two finger gripper as its end-effector. Within the simulation environment, a vision sensor is added between the gripper fingers to simulate the camera view. The VRep simulation is controlled via MATLAB through VRep's internal APIs. The aim of the interface is to communicate the \emph{FastOrient} orientation angle from MATLAB to VRep. 

Five different environments are simulated, replicating the different backgrounds tested. Five different objects with different shapes and colors are also selected for this (banana, orange, red bottle, blue mug and red tool in Figure \ref{experiment}.b). Each object is positioned in one of  the experiment backgrounds and for each of them, three different reaching approaches have been tested: Top, Lateral and Angulated. The final orientation of the gripper with respect to the object and the view from the vision sensor will confirm whether grasping is possible in the simulation and consequently, if the algorithm has worked correctly.

\subsection{Results and discussion}
\subsubsection{Target object selection performance} 
\begin{figure}[tp]
\includegraphics[width=\columnwidth]{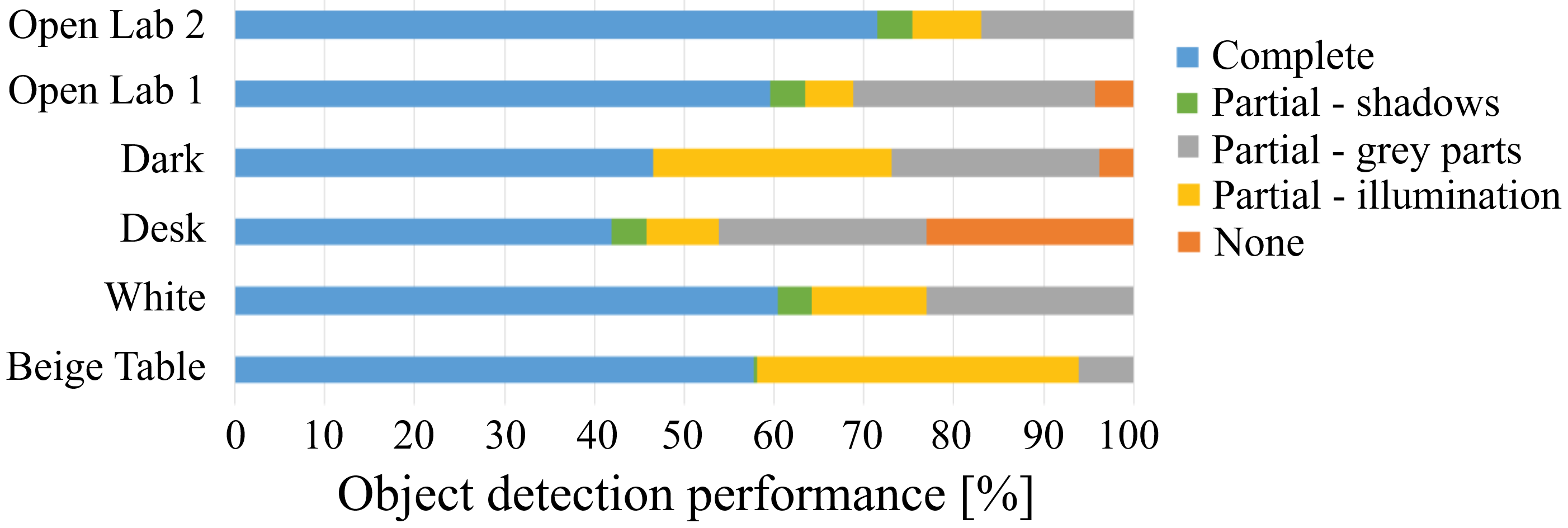}
\caption{Detailed object detection performance results in percentages of complete detection, partial detection due to shadows, illumination or grayscale parts in objects and non-detected items, for the six different experiment setups.}
\label{results_obj}
\end{figure}

Considering a general detection category which also includes partial detections, $94.8\%$ of cases are detected by \emph{FastOrient}. In particular, all items are detected with the Beige Table, White and Open Lab 1 background. $96\%$ of the objects are detected for the Dark and Open Lab 1 cases. A smaller percentage is detected for the Desk background ($77\%$) due to the presence of a large number of black pixels, which prevents the algorithm from painting grayscale objects in green and consequently affects their detection. Across all experiments and repetitions, $56.3\%$ of cases are completely detected. However, $2.63\%$, $16\%$ and $19.9\%$ of the detections are affected by shadows, illumination or the presence of grayscale parts in objects, respectively. $5.19\%$ of the objects are not detected at all. Full results can be seen in Figure \ref{results_obj}).
\subsubsection{Object orientation detection performance} Results show that all the complete detection cases result in complete orientations clearly aligned with the object's long axis. For the cases of partial detection affected by shadows, the majority of the cases give an aligned orientation where grasping is still allowed ($70\%$) while the others are defined as incorrectly oriented. Although the effect of illumination and the presence of grayscale parts may give a partial detection, the implication in the orientation retrieved from \emph{FastOrient} depends on the proportion of the area detected. Approximately $50\%$ of the cases affected by illumination still retrieve a correct orientation, $35\%$ result in slightly deviated orientations and $15\%$ have a wrong orientation. For the partial detection caused by grayscale parts, approximately $55\%$ of the cases result in a completely aligned orientation, $40\%$ are affected and give an orientation within the twenty degrees margin of error and $5\%$ of the cases result in a wrong orientation. Obviously, all the cases where the object is not detected lead to no orientation returned.
\subsubsection{Robot simulation performance} 
\begin{figure}[bp]
\includegraphics[width=\columnwidth]{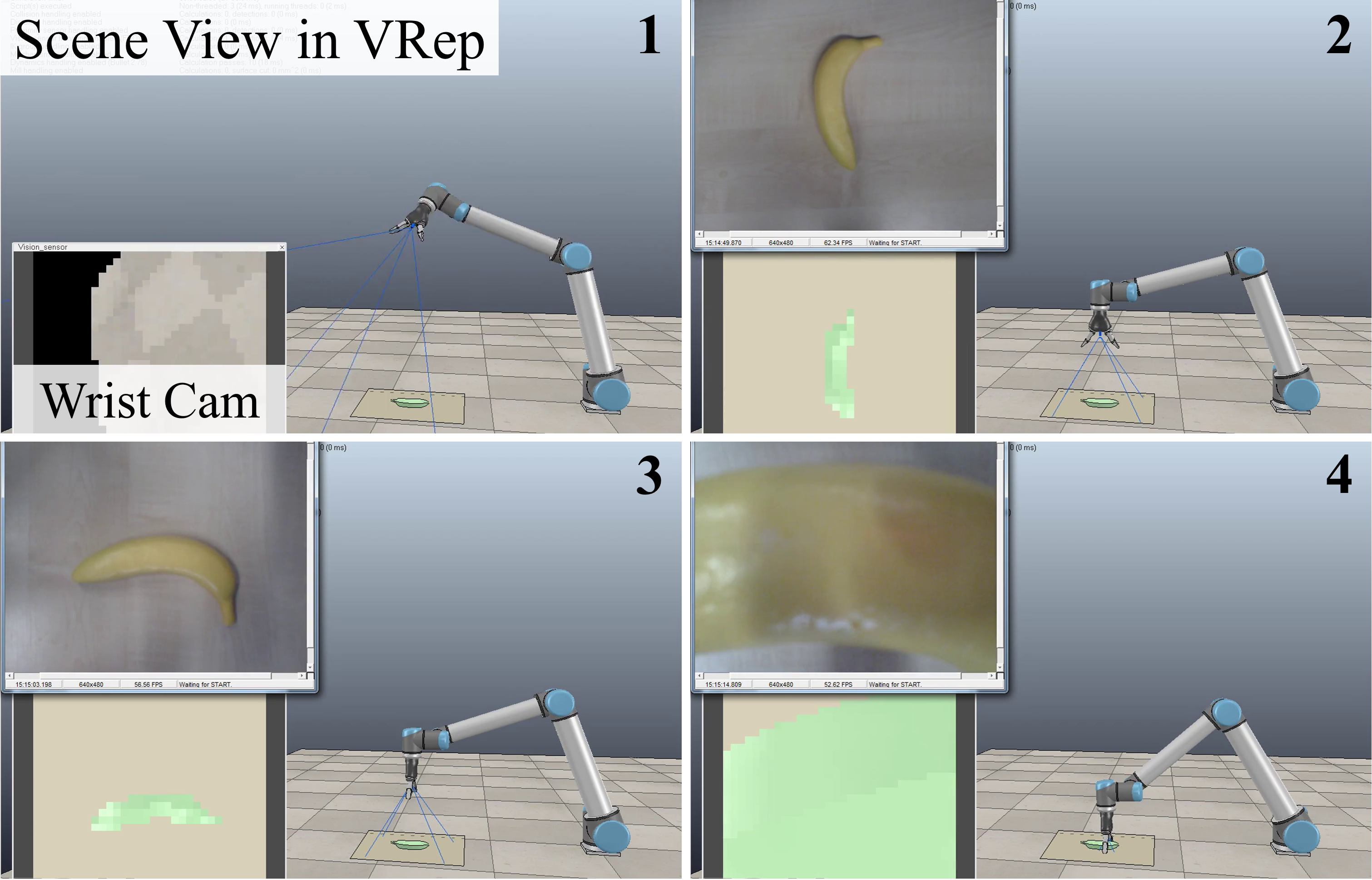}
\caption{Grasp orientation optimization algorithm in simulation. (1) The robot is reaching for the object location, (2) The vision sensor is used to detect the banana, its orientation and to then calculate the suitable grasp orientation, (3) The robot's wrist is rotated into the suitable grasp orientation, (4) The robot reaches in to fulfill the grasp.}
\label{results_robot}
\end{figure}

To test if the orientation of the gripper is the appropriate one to achieve the grasping of an object, simulations using the VRep environment and the UR10 robot are performed. In these simulations, the gripper is rotated  by \emph{FastOrient}. The final configuration of the gripper needs to have its fingers aligned with the yellow line (e.g. in Figure~\ref{orientation}.6). The robot's other joints are actuated so that the gripper gets close to the object to evaluate if the grasping is correct. The vision sensor added on the gripper allows to have visual feedback from the simulated robot movement. The example of a banana approached from the top is shown in Figure \ref{results_robot}, to visualize how the simulation works. During the approaching movement, the corresponding end-effector orientation is calculated by \emph{FastOrient}. This angle is sent to the VRep environment and the gripper is rotated (as shown in Figure \ref{results_robot}.3). Then, a faster approach takes place to position the gripper close to the object, Figure \ref{results_robot}.4. From this final configuration of the robot, the positioning of the gripper gives information of the grasping. If its fingers are perpendicular to the long axis of the object, as for the banana in Figure \ref{results_robot}.4, the object can be easily picked up.
\subsubsection{Repeatability} 
\begin{figure}[tp]
\includegraphics[width=\columnwidth]{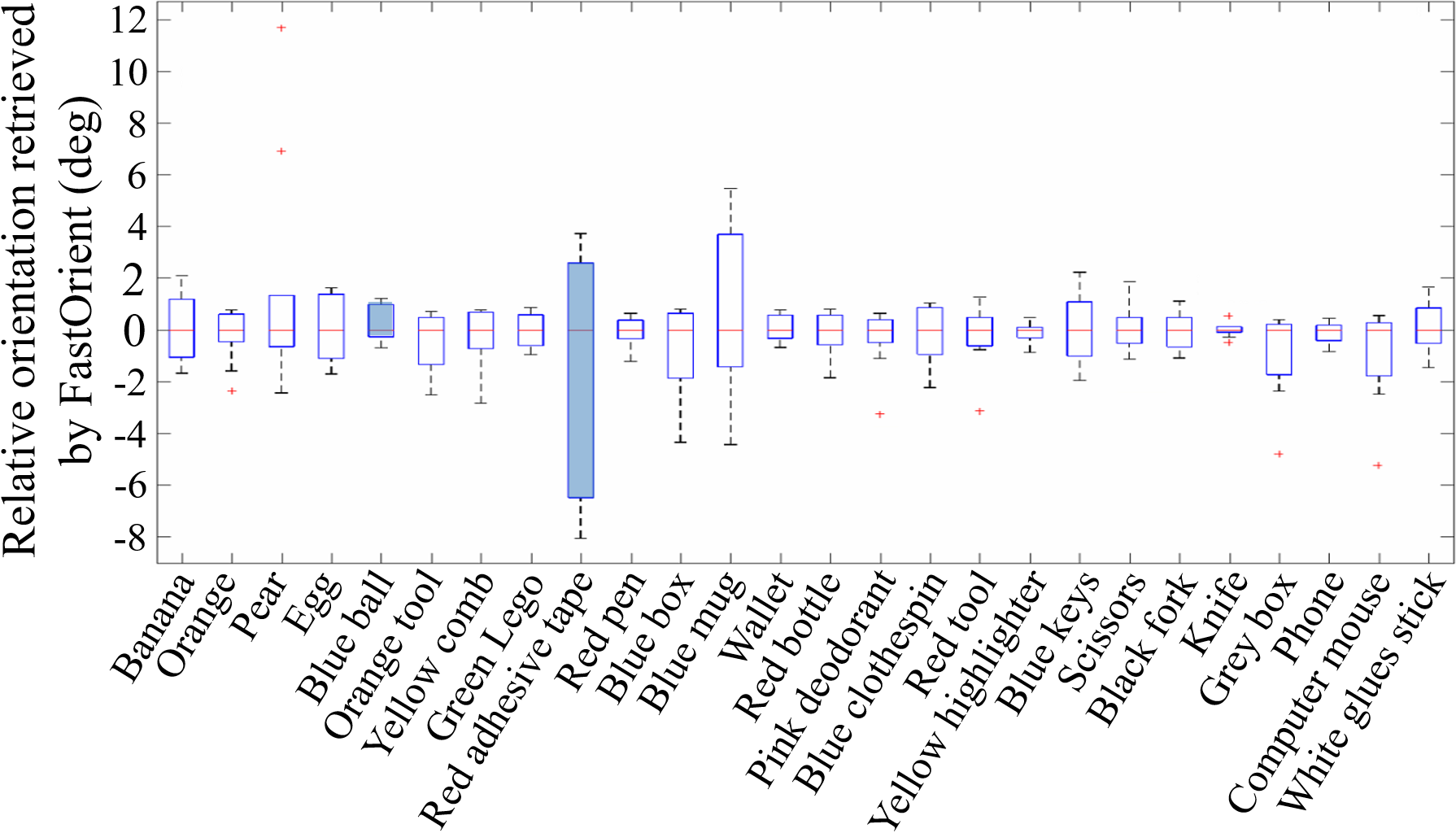}
\caption{Boxplot for each object tested in the Open Lab 2 background to evaluate the hand orientation algorithm for the ten repetitions. The boxplots painted in blue correspond to round objects.}
\label{results_repeat}
\end{figure}

Boxplots are drawn for each object and experiment to represent the variability of each orientation detection and consequently, study the reproducibility of the system. Although the object is not always positioned at orientation zero (horizontal line) in the setups, relative angles are computed so that all the variations are started from angle zero. The different computed orientations are set to be between \ang{-90} and \ang{90}. The performance of the system is precise and robust for all the experiments. It can be seen in the corresponding boxplots that, for most of the cases, the variability for the orientation given for the 10 repetitions is within a \ang{4} range or less for all experiments. This small variability can be due to small effects of illumination and shadows or to other effects such as the instability inherent to the approaching movement. For each experiment, the round objects' boxplots are the ones that show a higher variability between repetitions and have been marked in blue to differentiate them from the rest. This is because round objects do not have a clear orientation and hence, the algorithm's variability is justified. The boxplot for the "Open Lab 2" experiment setup is included as an example in Figure~\ref{results_repeat}.

Across our experiments, the variability for each experiment is very low, confirming the good reproducibility of the system. Nonetheless, we observe outliers. Experiments 3, 5 and 6 are the more accurate ones as their outliers are comprised in the \ang{-2} to \ang{2} range. For the rest of experiments, the outliers are spread further within a \ang{-4} to \ang{8} range. Note, that most objects can be grasped reliably with a larger misalignment than observed here. Accordingly, external effects such as the time of the day and lighting conditions in which the experiments are performed do not affect the performance of the algorithm in a considerable manner as experiments are performed independently at different times. Although during the early morning and late afternoon the effect of illumination can create more shadows in our data, yet our algorithm removes them properly as no extra variability is added to specific experiments.
\section{Conclusions}
We presented computer vision guided control of wrist orientation for wearable and assitive robotics to help individuals with  motor disabilities. Typically these devices  rely on the human user's direct control for the minutiae of interaction. In our case, we demonstrated that our \emph{FastOrient} algorithm works out-out-of-the-box which does not require a complex training learning pipeline and is sufficient to allow a natural wrist and grasp interaction. With the technology that we have demonstrated, we should be able to use gaze-based 3D positioning of the arm for grasping, as demonstrated in \cite{Tostado20163DActuators,Dziemian2016Gaze-basedDrawing,Maimon-Dror2017TowardsTracking}, wink-based detection of grasping intention \cite{Noronha2017WinkGloves}, and now automatic processing of the grasp orientation based on the findings of this paper, without the requirement for the user to fine-tune the orientation of the hand, resulting in a full, intuitive assistive system. We have demonstrated \emph{FastOrient} across 5 different realistic surfaces and for 26 different objects pertaining to standard activities of daily living. \emph{FastOrient} executed successful grasps in 91.1\% of the evaluated cases. While this method may not be suitable for objects with very complex shapes, it is nonetheless sufficient for the many typical objects used here. Our compact computer vision algorithm \emph{FastOrient} can be efficiently ported to small, light, low-power embedded systems, by cross-compiling from its MATLAB sources to numerous embedded platforms. This facilitates its deployment in  a variety of systems, such as upper arm prosthetics, orthotics or exoskeletons.
\addtolength{\textheight}{-12cm}   


\bibliographystyle{ieeetr}
\bibliography{Mendeley.bib,additional.bib}

\end{document}